\documentclass[runningheads]{llncs}
\usepackage[T1]{fontenc}
\usepackage{graphicx,verbatim}
\usepackage{amsmath,amssymb}
\usepackage{multirow}
\usepackage{booktabs}
\usepackage{color}
\begin{document}
\title{ReMeDI: Refined Memory for Disambiguation of Identities with SAM3 in Surgical Segmentation}
\titlerunning{ReMeDI-SAM3}
%

\author{Valay Bundele,
Mehran Hosseinzadeh,
Hendrik P.A. Lensch}
\authorrunning{Valay Bundele,
Mehran Hosseinzadeh, Hendrik P.A. Lensch}
%
\institute{
University of Tübingen, Germany\\
\email{\{valay.bundele, mehran.hosseinzadeh, hendrik.lensch\}@uni-tuebingen.de}
  }
\maketitle              
\begin{abstract}
Accurate surgical instrument segmentation in endoscopy is crucial for computer-assisted interventions, yet remains challenging due to frequent occlusions, rapid motion, and long-term instrument re-entry. While SAM3 provides a powerful spatio-temporal framework for video object segmentation, its performance in surgical scenes is limited by indiscriminate memory updates, fixed memory capacity, and weak identity recovery after occlusions. We propose ReMeDI-SAM3, a training-free extension of SAM3, that addresses these limitations through three components: (i) relevance-aware memory filtering with a dedicated occlusion-aware memory for storing pre-occlusion frames, (ii) a piecewise interpolation scheme that expands effective memory capacity, and (iii) a feature-based re-identification module with temporal voting for reliable post-occlusion identity disambiguation. Together, these components mitigate error accumulation and enable reliable recovery after occlusions. Evaluations on EndoVis17, EndoVis18 and CholecSeg8k under a zero-shot setting show mcIoU improvements of around 5.8\%, 8\%, and 2\% respectively, over vanilla SAM3, outperforming even prior training-based approaches.

\keywords{Video Object Segmentation, Surgical Video Analysis}

\end{abstract}
\section{Introduction}
Surgical instrument segmentation is central to computer-assisted interventions, supporting tasks such as tracking, workflow analysis, and intraoperative guidance \cite{gonzalez2020isinet,yuan2025temporal,zhou2023text}. However, surgical videos feature long unstructured sequences, frequent occlusions and re-entry, making long-term temporal consistency and identity preservation challenging. Based on the recently introduced general-purpose SAM3~\cite{carion2025sam}, we propose several training-free extensions to address these issues for reliable zero-shot surgical instrument tracking.

With the rise of foundation models, SAM \cite{kirillov2023segment} enabled zero-shot prompt-based segmentation, but its direct use in surgical videos is unreliable due to domain shift and prompt dependency \cite{yu2024adapting}. SAM~2 \cite{ravi2024sam} introduced spatio-temporal memory for video segmentation, inspiring extensions such as SAMURAI \cite{yang2024samurai}, and DAM4SAM \cite{videnovic2025distractor}, as well as surgical adaptations including MA-SAM2 \cite{yin2025memory}, SAMed-2 \cite{yan2025samed} and SurgSAM2 \cite{liu2024surgical}. Despite these advances, extended occlusions, frequent re-entry, and large viewpoint changes still cause identity failures. The recently introduced SAM3 unifies open-vocabulary detection, segmentation, and tracking through spatio-temporal memory, providing a strong baseline for video object segmentation. However, its reliability-agnostic memory update allows low-quality predictions to be written into memory. In surgical videos, where occlusions and visual artifacts are frequent, this causes error accumulation and identity drift, especially after prolonged occlusions and viewpoint changes. While stricter confidence thresholds can mitigate memory contamination, they may also discard low-visibility but identity-critical pre-occlusion frames, exposing a trade-off between temporal stability and reliable identity recovery.

To address these issues, we propose a dual-partitioned memory design. First, a \emph{relevance-aware memory partition} selectively stores high-confidence frames to prevent memory contamination. Second, an \emph{occlusion-aware memory partition} retains pre-occlusion frames under relaxed criteria to preserve identity-critical cues for recovery. As retaining lower-confidence frames increases the risk of identity drift, we further introduce a \emph{feature-based re-identification module} that verifies and corrects recovered identities using multi-scale appearance descriptors, combined with \emph{temporal voting} for robust disambiguation across frames.
Long surgical procedures additionally require long-horizon temporal context, yet SAM3 is constrained by a fixed set of temporal positional encodings, which limits its effective memory capacity and causes informative early frames to be overwritten. We address this with a \emph{novel memory expansion scheme} based on piecewise interpolation of temporal positional encodings, enabling larger memory without retraining. Together, these components form a training-free extension of SAM3 that significantly improves occlusion robustness, long-term tracking stability, and reliable instrument re-identification in videos. To the best of our knowledge, this is the first SAM-based extension that explicitly targets both accurate re-identification and scalable memory. 
Our main contributions are: 
\begin{itemize}
\item We introduce a dual-memory design that combines relevance-aware propagation with a dedicated occlusion–aware memory for post-occlusion recovery. 
\item We incorporate a feature-based re-identification module with temporal voting for explicit identity verification and correction after occlusions.
\item We propose a novel memory expansion strategy enabling long-horizon memory retention without any retraining. Overall, our approach shows zero-shot mcIoU gains of $5.8\%$, $8\%$ and $2\%$ on EndoVis17, EndoVis18 and CholecSeg8k over vanilla SAM3, while also outperforming recent training-based methods.
\end{itemize}

\section{Methodology}
\subsection{Preliminaries}
Let $\mathcal{V}=\{I_1,\dots,I_T\}$ be an endoscopic video, where each frame $I_t\in\mathbb{R}^{H\times W\times 3}$. The task is to predict class-level segmentation maps, $\mathcal{S}_t=\{P_t^{(i)}\}_{i=1}^N$ for each frame, where $P_t^{(i)}\in\{0,1\}^{H\times W}$ is the binary mask of instrument $i$, and $N$ is the number of instruments.
SAM3 is a video object segmentation model that extends SAM2 with text-promptable segmentation. Given an object prompt (text, points, boxes, or masks) on a reference frame $t_0$, it propagates the object state to predict masks $\{P_t\}_{t=t_0}^{T}$. Its architecture comprises an image encoder, a detector, a prompt encoder, a memory bank, and a mask decoder. The image encoder and detector share a unified vision backbone \cite{bolya2025perception}, and the detector follows a DETR-based open-vocabulary design \cite{carion2020end}. Sparse prompts are embedded by the prompt encoder to condition the mask decoder. The mask-conditioned features from the prompted frame and the six most recent frames are stored in memory. Current-frame features attend to this memory to produce temporally consistent masks. To handle ambiguity, SAM3 predicts multiple candidate masks with confidence scores and selects the highest-quality one, updating memory in a FIFO manner. While effective in general domains, this indiscriminate memory update limits long-term consistency under severe occlusions and frequent re-entry in surgery.

\begin{figure}[t]
    \centering
    \includegraphics[width=0.9\textwidth]{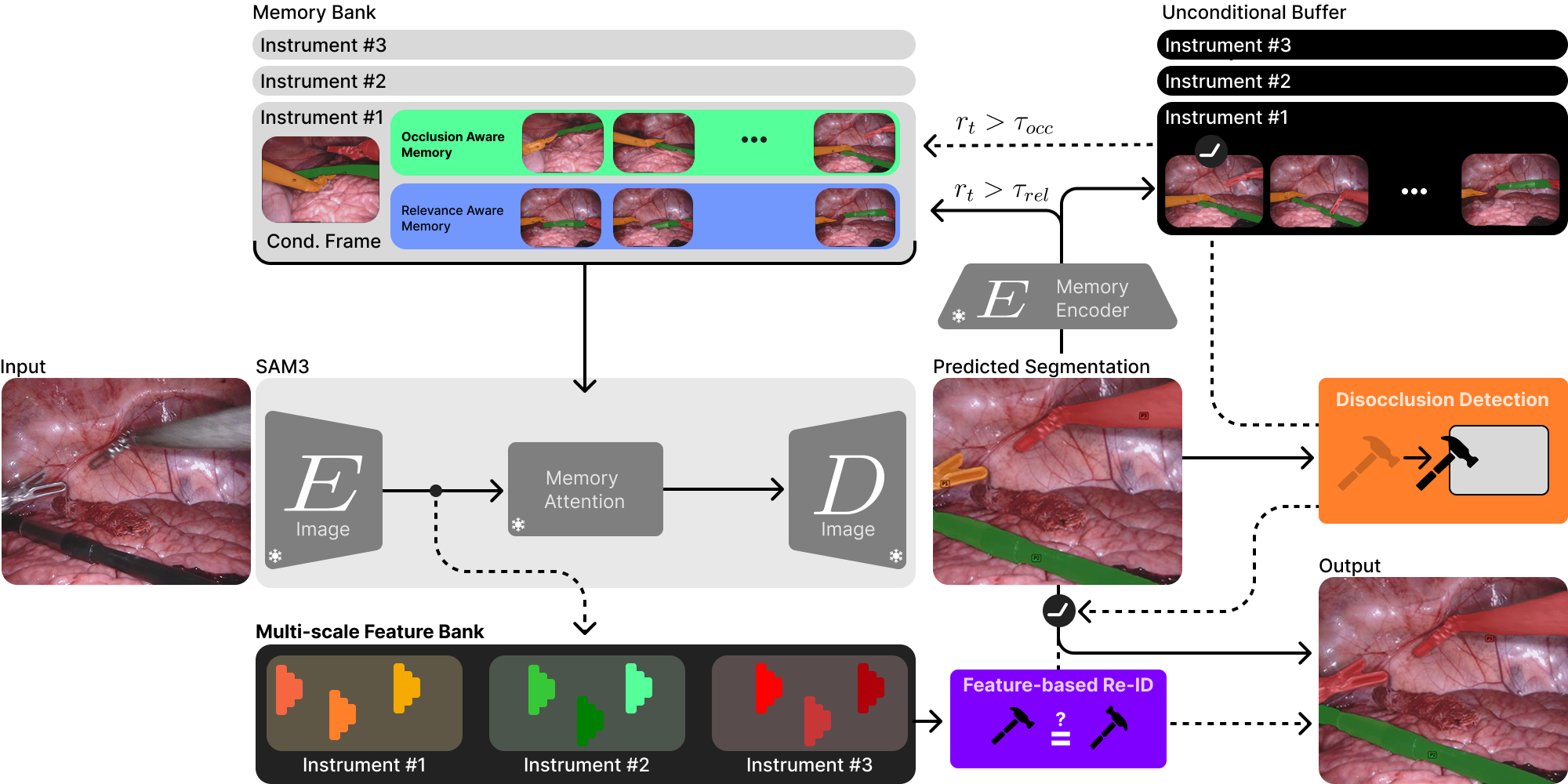}
    \caption{ReMeDI-SAM3. We extend SAM3 with a dual-memory design and a feature-based ReID module. For each instrument, the memory is divided into a \emph{relevance-aware memory} that stores high-confidence entries and an \emph{occlusion-aware memory} that is populated upon disocclusion using lower-confidence pre-occlusion frames drawn from an Unconditional Buffer that stores all past frames. When disocclusion is detected (tool reappears), occlusion-aware memory is first updated, after which feature-based ReID module verifies or reassigns predicted identity using a multi-scale feature bank.}

    \label{fig:pipeline}
\end{figure}
\subsection{ReMeDI-SAM3}

We propose \textbf{ReMeDI-SAM3}: \textbf{Re}fined \textbf{Me}mory for \textbf{D}isambiguation of \textbf{I}dentities with \textbf{SAM3}, a training-free extension of SAM3 to enhance temporal consistency and identity preservation in surgical videos. The pipeline is shown in Figure~\ref{fig:pipeline}. Our approach (1) restructures SAM3 memory into two components: (i) a \emph{relevance-aware memory} that admits only high-confidence frames, and (ii) an \emph{occlusion–aware memory} that selectively retains pre-occlusion appearance cues. Building upon this design, we further introduce (2) a \emph{novel memory expansion scheme}, and (3) a \emph{feature-based re-identification module with temporal voting} for robust post-occlusion identity verification and correction. 

\subsubsection{Relevance-Aware and Occlusion-Aware Memory} \label{sec:memories}

In SAM3, a fixed-size memory bank retains recent frames regardless of prediction reliability, which in surgical videos often introduces noisy masks and causes error accumulation. To mitigate this, we split the total memory of size $M$ into two components: a \emph{relevance-aware memory} for stable long-term tracking and an \emph{occlusion-aware memory} for post-occlusion recovery, each allocated $M/2$ slots.

\textbf{Relevance-Aware.} For each predicted mask $P_t$, SAM3 outputs a quality score $c_t$ and an objectness score $s_t$, from which a reliability score is defined as, $r_t = s_t \cdot c_t$.
The relevance-aware memory consists of recent frames whose reliability exceeds a threshold $\tau_{\text{rel}}$:
\[
\mathcal{U}_{\text{rel}} = \operatorname{Top}_{M/2}\!\left(\{ I_t \mid r_t \ge \tau_{\text{rel}} \}\right),
\qquad |\mathcal{U}_{\text{rel}}| \le \frac{M}{2},
\]
where $\operatorname{Top}_{M/2}(\cdot)$ denotes selecting the temporally most recent $M/2$ frames. This gating limits memory updates to reliable frames, stabilizing propagation.

\textbf{Occlusion-Aware.} Just before occlusion, instruments often exhibit reduced visibility and thus lower reliability scores, despite carrying critical identity cues for re-identification. To preserve this information, we maintain an \emph{unconditional buffer} $\mathcal{U}$ that stores all past frames irrespective of $r_t$. Upon detecting an occlusion recovery event (i.e., when $s_t$ transitions from zero to positive), we populate the occlusion-aware memory by selecting the $M/2$ most recent frames from the unconditional buffer that satisfy a relaxed reliability constraint:
\[
\mathcal{U}_{\text{occ}} =
\operatorname{Top}_{M/2}\!\left(\{ I_t \in \mathcal{U} \mid r_t \ge \tau_{\text{occ}} \}\right),
\qquad |\mathcal{U}_{\text{occ}}| \le \frac{M}{2},
\]
where $\tau_{\text{occ}} < \tau_{\text{rel}}$. This helps preserve identity-discriminative appearance cues.
\begin{figure}[t]
    \centering
    \includegraphics[width=0.85\textwidth]{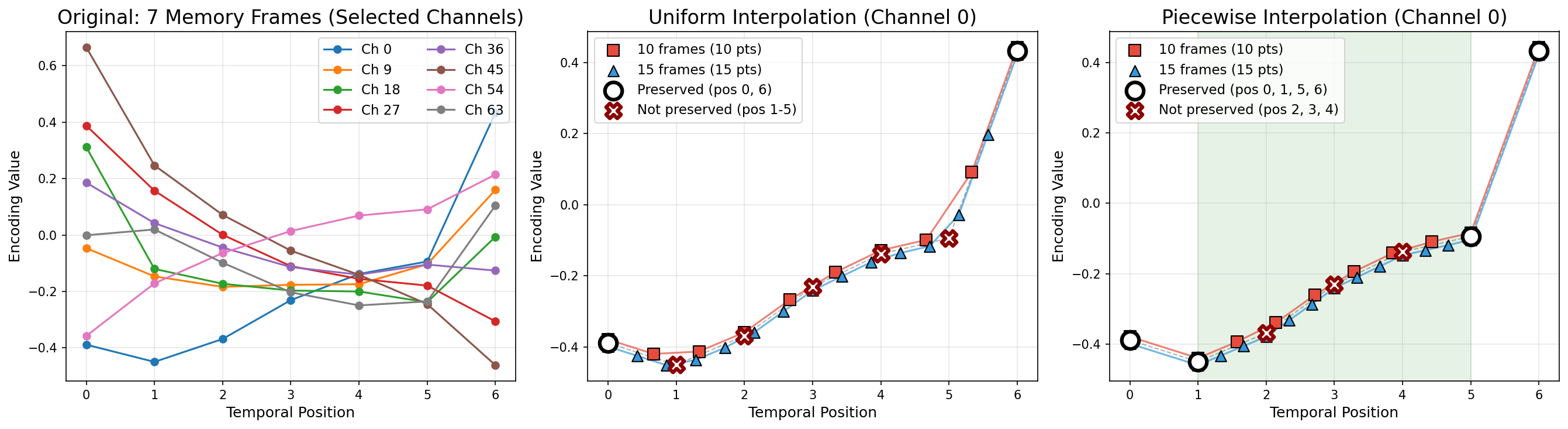}
    \caption{Visualization of temporal positional encodings and memory expansion strategies. Left: select channels of original temporal positional embeddings. Mid: uniform interpolation distributes new positions evenly. Right: piecewise interpolation preserves boundary embeddings and samples new positions only in interior region.}
\label{fig:pos_encoding}
\end{figure}
\subsubsection{Memory Capacity Expansion} \label{sec:interpolation}

Although SAM3 supports a configurable memory size, it uses only $M{=}7$ fixed temporal positional embeddings, limiting reliable indexing for long videos. This is problematic in surgical scenarios with severe occlusions and large appearance changes. Let $\{\mathbf{p}_0,\dots,\mathbf{p}_6\}$ denote the original embeddings. As shown in Figure~\ref{fig:pos_encoding}, the boundary segments $[0,1]$ and $[5,6]$ differ from the interior $[1,5]$, indicating stronger temporal priors at the extremes. We therefore keep the boundary segments unchanged and interpolate only within the interior (see Figure~\ref{fig:pos_encoding} (right)).
For a target memory size $M>7$, we fix the boundary encodings as
$\tilde{\mathbf{p}}_0=\mathbf{p}_0$ and $\tilde{\mathbf{p}}_{M-1}=\mathbf{p}_6$.
The remaining $M-2$ encodings are obtained by linearly resampling the interior sequence $(\mathbf{p}_1,\dots,\mathbf{p}_5)$ to $M-2$ uniformly spaced positions. Specifically, for $k\in\{1,\dots,M-2\}$,
\[
t_k=\frac{k-1}{M-3}, \qquad
u_k = 1 + 4t_k,
\]
\[
\tilde{\mathbf{p}}_k
= (1-\alpha_k)\mathbf{p}_{\lfloor u_k \rfloor}
  + \alpha_k \mathbf{p}_{\lceil u_k \rceil},
\quad
\alpha_k = u_k - \lfloor u_k \rfloor.
\]
This corresponds to linear interpolation of $(\mathbf{p}_1,\dots,\mathbf{p}_5)$ with aligned endpoints. The resulting encodings preserve boundary semantics while enabling denser temporal indexing in the interior region, effectively expanding the memory capacity.

\subsubsection{Feature-Based Re-Identification} \label{sec:re-id}

Despite improved memory, long occlusions can still cause identity drift at disocclusions, particularly because the occlusion-aware memory admits lower-confidence pre-occlusion frames to preserve identity cues. To address this, we introduce a feature-based re-identification (ReID) module that validates and corrects identity upon recovery using appearance descriptors, aggregating predictions over a temporal window to ensure robustness.

For each instrument class $i$, we maintain a feature bank $\mathcal{B}^i$ constructed from selected frames observed up to time $t$. A frame contributes to $\mathcal{B}^i$ only if (i) it has a high reliability score $r_t$, and (ii) its prediction certainty is high, measured by bounding-box IoU agreement among three candidate masks. From each valid frame, we extract multi-scale appearance features by averaging the backbone feature map within the predicted mask:
$
\mathbf{f}^{i}_{t,l}
=
\frac{1}{|M^i_t|}
\sum_{x \in M^i_t}
\mathbf{F}_{t,l}(x),
$
where $\mathbf{F}_{t,l}$ is the backbone feature map at scale $l$ and $M^i_t$ is the predicted mask. The resulting multi-scale descriptors $\{\mathbf{f}^{i}_{t,l}\}_l$ are appended to $\mathcal{B}^i$ and updated online.

An occlusion is detected when the objectness score drops to zero. When a non-zero score reappears, a recovery phase is triggered and identity is verified over the next $K$ frames. Let $i$ be the predicted class at recovery. We compute self-similarity to $\mathcal{B}^i$ and cross-class similarity to banks of other classes $\mathcal{B}^j$ using cosine similarity aggregated across scales and averaged over the $K$-frame window:
$
s^{\text{self}} = \frac{1}{K}\sum_t s_t^{\text{self}}, 
s^{\text{other}} = \frac{1}{K}\sum_t s_t^{\text{other}}.
$
The identity is accepted if $s^{\text{self}} \ge s^{\text{other}}$. If another class yields higher similarity, the label is reassigned to that class.

\subsubsection{Final Inference Pipeline}
Each instrument is initialized from its first visible mask and tracked independently with a dedicated memory. Memory updates are regulated by relevance-aware filtering. Upon reappearance after occlusion, the occlusion-aware memory is updated using pre-occlusion frames with a lower confidence threshold, followed by prediction and feature-based re-identification for identity verification and correction. Finally, predictions from all instruments are fused via quality-weighted mask fusion to obtain the final segmentation map.

\section{Experiments}
\subsection{Datasets and Evaluation Metrics}
We evaluate on three public benchmarks, EndoVis2017 \cite{allan20192017}, EndoVis2018 \cite{allan20202018}, and CholecSeg8k \cite{hong2020cholecseg8k}. EndoVis2017 contains eight annotated sequences (225 frames each), pre-processed following \cite{shvets2018automatic}; we evaluate on all sequences and compare against prior training-based 4-fold cross-validation results. EndoVis2018 includes 11 training and 4 validation videos (149 frames each); we report performance on validation set for fair comparison. Both datasets provide annotations for seven instrument categories.
CholecSeg8k is a surgical segmentation dataset with 8080 frames for multiple surgical structures (13 different categories). For evaluation, we report Challenge IoU (cIoU) \cite{allan20192017}, which computes IoU only for instruments present in each frame, along with IoU and mean class IoU (mcIoU).

\subsection{Implementation details}
We set $\tau_{rel}=0.95$ and $\tau_{occ}=0.75$, and perform re-identification over a window of $K=3$ frames. Hyperparameters are selected on the EndoVis18 train set and kept fixed across datasets. All experiments are conducted on an RTX 4090 GPU.

\subsection{Results}
\begin{table*}[t]
\centering
\caption{Quantitative comparison on the EndoVis17 and EndoVis18 datasets (\%). The best results are in \textbf{bold}, second-best are \underline{underlined}.}
\resizebox{\textwidth}{!}{%
\begin{tabular}{c| l l c c c c c c c c c c c c}
 & \multirow{2}{*}{Category} & \multirow{2}{*}{Method} & \multirow{2}{*}{Challenge IoU} & \multirow{2}{*}{IoU} & \multirow{2}{*}{mcIoU} & \multicolumn{9}{c}{{Instrument Categories}} \\
\cline{7-15}
 &  &  &  &  &  & BF & PF & LND & MCS & UP & VS & GR & SI & CA \\
\cline{1-15}

\multirow{13}{*}{\rotatebox{90}{\textbf{EndoVis17}}}
& \multirow{4}{*}{Specialist}
& ISINet  & 55.62 & 52.20 & 28.96 & 38.70 & 38.50 & 50.09 & 28.72 & 12.56 & 27.43 & 2.10 & - & - \\
& & S3Net  & 72.54 & 71.99 & 46.55 & \textbf{75.08} & 54.32 & 61.84 & 43.23 & 28.38 & 35.50 & 27.47 & - & - \\
& & MATIS Frame  & 68.79 & 62.74 & 37.30 & 66.18 & 50.99 & 52.23 & 19.27 & 23.90 & 32.84 & 15.71 & - & - \\
& & TP-SIS & 63.37 & 63.37 & 52.74 & 66.42 & 45.46 & 75.20 & 44.02 & 34.67 & 73.44 & 29.95 & - & - \\
\cline{2-15}

& \multirow{9}{*}{SAM-based}
& TrackAnything & 67.41 & 64.50 & 62.97 & 55.42 & 44.46 & 62.43 & 67.03 & 65.17 & \textbf{83.68} & \underline{62.59} & - & - \\
& & PerSAM (Zero-Shot) & 42.47 & 42.47 & 41.80 & 53.99 & 25.89 & 50.17 & 47.33 & 38.16 & 52.87 & 24.24 & - & - \\
& & SurgicalSAM & 69.94 & 69.94 & 67.03 & 68.30 & 51.77 & 75.52 & \textbf{86.95} & 60.80 & 68.24 & 57.63 & - & - \\
& & SP-SAM & 73.94 & \underline{73.94} & \underline{71.06} & 68.89 & 53.16 & \textbf{83.80} & \underline{84.91} & 61.05 & 73.20 & \textbf{72.40} & - & - \\
& & MA-SAM2 (Zero-Shot) & 62.49 & 62.49 & 59.89 & 54.41 & 50.41 & 64.73 & 72.64 & 70.85 & 73.72 & 32.66 & - & - \\
& & SAM3 (Mask, Zero-Shot) & \underline{78.08} & 70.76 & 68.42 & 55.62 & \underline{65.30} & 76.33 & 75.32 & \textbf{90.09} & \underline{80.51} & 38.36 & - & - \\
& & ReMeDI-SAM3 (Ours) & \textbf{81.34} & \textbf{76.65} & \textbf{74.29} &\underline{69.00} &\textbf{68.97} &\underline{77.14} &78.82 &\underline{89.96} &79.99 &56.15 &- &- \\
\midrule

\multirow{10}{*}{\rotatebox{90}{\textbf{EndoVis18}}}
& \multirow{4}{*}{Specialist}
& ISINet & 73.03 & 70.94 & 40.21 & 73.83 & 48.61 & 30.98 & 88.16 & 2.16 & - & - & 37.68 & 0.00 \\
& & S3Net & 75.81 & 74.02 & 42.58 & 77.22 & 50.87 & 19.83 & \underline{92.12} & 7.44 & - & - & 50.59 & 0.00 \\
& & MATIS Frame & 82.37 & 77.01 & 48.65 & 83.35 & 38.82 & 40.19 & \textbf{93.18} & 16.17 & - & - & 64.49 & 4.32 \\
& & TP-SIS & 84.92 & 83.61 & 65.44 & 84.28 & \underline{73.18} & 78.88 & 66.67 & 39.12 & - & - & \textbf{92.20} & 23.73 \\
\cline{2-15}

& \multirow{6}{*}{SAM-based}
& TrackAnything & 65.72 & 60.88 & 38.60 & 72.90 & 31.07 & 64.73 & 61.05 & 17.93 & - & - & 10.24 & 12.28 \\
& & PerSAM (Zero-Shot) & 49.21 & 49.21 & 34.55 & 51.26 & 34.40 & 46.75 & 52.28 & 25.62 & - & - & 16.45 & 15.07 \\
& & SurgicalSAM & 80.33 & 80.33 & 58.87 & 83.66 & 65.63 & 58.75 & 88.56 & 21.23 & - & - & 54.48 & \underline{39.78} \\
& & SP-SAM & 84.24 & \underline{84.24} & 65.71 & \underline{87.60} & 65.07 & 61.95 & 92.08 & 34.99 & - & - & 58.30 & \textbf{59.96} \\
& & SAM3 (Mask, Zero-Shot) & \underline{88.04} & 81.82 & \underline{66.46} & 82.25 & \textbf{74.92} & \textbf{88.14} & 88.38 & \underline{44.78} & - & - & 53.75 & 32.96 \\
& & ReMeDI-SAM3 (Ours) & \textbf{88.10} & \textbf{85.34} & \textbf{74.37} &\textbf{87.73} &72.65 &\underline{86.60} &88.80 &\textbf{80.44} &- &-  &\underline{65.38} &39.01

\label{ev1718results}
\end{tabular}%
}
\end{table*}
\textbf{Quantitative Comparison.}
We evaluate our method on EndoVis17, EndoVis18, and CholecSeg8k, comparing against specialist surgical segmentation models and recent SAM-based approaches, including ISINet~\cite{gonzalez2020isinet}, S3Net~\cite{baby2023forks}, MATIS~\cite{ayobi2023matis}, TP-SIS~\cite{zhou2023text}, TrackAnything~\cite{yang2023track}, PerSAM~\cite{zhang2023personalize}, SurgicalSAM~\cite{yue2024surgicalsam}, SP-SAM~\cite{yue2023surgicalpart}, MA-SAM2~\cite{yin2025memory}, and vanilla SAM3. All results of our method are obtained in a fully training-free setting, and we report both global and class-wise IoUs.

Table~\ref{ev1718results} summarizes the results. ReMeDI-SAM3 consistently outperforms vanilla SAM3 across all benchmarks. On EndoVis17, we achieve improvements of 6\% IoU and 5.8\% mcIoU. On EndoVis18, we obtain gains of 3.5\% IoU and 8\% mcIoU. 
The larger mcIoU gain reflects improved suppression of false positives for absent instruments via our identity-aware design. For example, in Sequence~5 of EndoVis18, SAM3 misclassifies a returning Prograsp Forceps (PF) as Ultrasound Probe (UP), whereas our method correctly restores PF and suppresses UP, yielding a 36\% IoU improvement for UP. Although not best in every individual category, our method achieves consistently strong performance across classes, leading to superior overall metrics. 
On CholecSeg8k, a broader surgical semantic segmentation benchmark, we observe consistent gains of 1\% IoU and 2\% mcIoU as shown in Table~\ref{cholecseg}. Overall, ReMeDI-SAM3 surpasses all zero-shot baselines and even outperforms several training-based approaches (e.g., SurgicalSAM, SP-SAM), demonstrating the effectiveness and robustness of our design.

\begin{figure}[t]
    \centering
    \includegraphics[width=0.35\textwidth, angle=90]{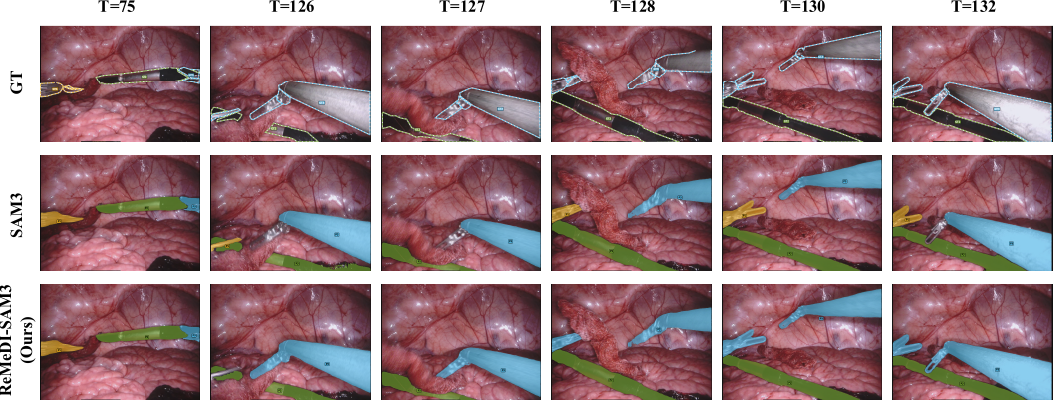}
\caption{ReMeDI-SAM3 vs SAM3. ReMeDI-SAM3 correctly detects the second blue instance entering later while SAM3 confuses it with the occluded yellow instrument.}
    \label{fig:ev17_1}
\end{figure}

\textbf{Qualitative Comparison.}
Figure~\ref{fig:ev17_1} illustrates a challenging occlusion case on EndoVis17. The yellow instrument (Bipolar Forceps) exits at $T{=}75$, and a second blue instrument (Prograsp Forceps) enters later ($T{=}126$--$132$). ReMeDI-SAM3 briefly misses the new instrument at $T{=}126$ but correctly assigns the blue identity once sufficient evidence accumulates. In contrast, SAM3 preserves the old yellow identity after occlusion. This shows that ReMeDI-SAM3 reliably recovers correct identities even after instrument turnover, highlighting the effectiveness of our method in maintaining identity consistency under occlusions.

\subsection{Ablation Studies}
We analyze the impact of memory size, temporal interpolation, and components of ReMeDI-SAM3: relevance-aware memory (RM), occlusion-aware memory (OM), memory expansion (ME), and feature-based re-identification (ReID).

\textbf{Model Components.} 
Table~\ref{ind_components} analyzes the individual and combined contributions of different model components on EndoVis17. Relevance-aware memory filtering provides a 3.5\% mcIoU gain over SAM3 by suppressing noisy updates. Adding the occlusion-aware memory partition yields a further 0.5\% mcIoU. The feature-based re-identification module contributes 1.4\% IoU by correcting post-occlusion identity drift, while expanding memory adds 0.8\% IoU. Overall, the final model achieves a total improvement of 6.0\% IoU and 5.9\% mcIoU, confirming that the components are jointly essential for robust long-horizon segmentation.

\textbf{Memory Expansion.} Expanding memory with our proposed interpolation strategy yields clear performance gains on EndoVis17 (Table~\ref{expansion}). Increasing the memory to 15 frames improves all metrics by around 0.8\% , indicating the benefit of additional temporal context. Further expansion slightly degrades performance, suggesting that overly large memory might introduce less informative context.

\textbf{Interpolation Scheme.} 
Replacing the proposed piecewise interpolation with uniform interpolation for memory expansion causes a drop of about $1.2\%$  cIoU and $1\%$ IoU on EndoVis17 (Table~\ref{interpolation}). Uniform resampling distorts learned boundary temporal priors whereas piecewise interpolation preserves the semantic roles of the earliest and latest memory positions while densifying the interior. 

\begin{table}[t]
\centering
\begin{minipage}{0.48\columnwidth}
\centering
\caption{Effect of Memory Expansion (\%) on SAM3 + RM + OM + ReID.}
\label{tab:memory_expansion}
\resizebox{\columnwidth}{!}{
\begin{tabular}{c c c c}
\toprule
Memory Size & Challenge IoU & IoU & mcIoU \\
\midrule
7  & 80.58 & 75.84 & 73.59 \\
15 & \textbf{81.34} & \textbf{76.65} & \textbf{74.29} \\
20 & 80.77 & 76.11 & 73.57 \\
\bottomrule
\end{tabular}
\label{expansion}
}
\end{minipage}
\hfill
\begin{minipage}{0.48\columnwidth}
\centering
\caption{Uniform vs.\ Piecewise Temporal Positional Encoding on ReMeDI-SAM3 (M=15).}
\label{tab:uniform_vs_piecewise}
\resizebox{\columnwidth}{!}{
\begin{tabular}{l c c c}
\toprule
Method & Challenge IoU & IoU & mcIoU \\
\midrule
Uniform   & 80.13 & 75.66 & 74.20 \\
Piecewise & \textbf{81.34} & \textbf{76.65} & \textbf{74.29} \\
\bottomrule
\end{tabular}
\label{interpolation}
}
\end{minipage}
\end{table}

\begin{table}[t]
\centering
\begin{minipage}{0.47\columnwidth}
\centering
\caption{Impact of individual components of ReMeDI-SAM3 on EndoVis17.}
\label{tab:ind_components}
\resizebox{\columnwidth}{!}{
\begin{tabular}{l c c c}
\toprule
Method & Challenge IoU & IoU & mcIoU \\
\midrule
Vanilla SAM3 
& 78.08 & 70.76 & 68.42 \\
+ RM
& 79.94 & 73.74 & 72.03 \\
+ RM + OM
& 80.93 & 74.46 & 72.53 \\
+ RM + OM + ReID 
& 80.58 & 75.84 & 73.58 \\
+ RM + OM + ReID + ME 
& \textbf{81.34} & \textbf{76.65} & \textbf{74.29} \\
\bottomrule
\label{ind_components}
\end{tabular}
}
\end{minipage}
\hfill
\begin{minipage}{0.5\columnwidth}
\centering
\caption{Zero-shot performance comparison on CholecSeg8k.}
\label{tab:cholecseg}
\resizebox{\columnwidth}{!}{
\begin{tabular}{l c c c}
\toprule
Method & Challenge IoU & IoU & mcIoU \\
\midrule
SAM3 (Mask)   & 88.33 & 87.81 & 79.37 \\
ReMeDI-SAM3 (Ours) & \textbf{89.68} & \textbf{89.07} & \textbf{81.25} \\
\bottomrule
\end{tabular}
\label{cholecseg}
}
\end{minipage}
\end{table}


\section{Conclusion}
We present ReMeDI-SAM3, a training-free extension of SAM3 for robust surgical instrument segmentation. Our framework integrates relevance-aware memory filtering for stable long-term propagation, occlusion-aware memory and feature-based re-identification for reliable post-occlusion recovery, and a memory expansion strategy for long-horizon reasoning. Extensive zero-shot evaluations on EndoVis17, EndoVis18 and CholecSeg8k show consistent improvements over vanilla SAM3 and prior methods, particularly in mcIoU, demonstrating superior identity preservation and false-positive suppression.  

\section{Acknowledgments}
The work described in this paper was conducted in the framework of the Graduate School 2543/1 “Intraoperative Multi-Sensory Tissue Differentiation in Oncology" (project ID 40947457) funded by the German Research Foundation (DFG - Deutsche Forschungsgemeinschaft). This work has been supported by the Deutsche Forschungsgemeinschaft (DFG) – EXC number 2064/1 – Project number 390727645. The authors thank the International Max Planck Research School for Intelligent Systems (IMPRS-IS) for supporting Valay Bundele and Mehran Hosseinzadeh.
%
%
%
\bibliographystyle{splncs04}
\bibliography{bibliography}
\end{document}